\documentclass{article}

\renewcommand{\thetable}{\Roman{table}}
\usepackage{spconf,amsmath,graphicx, multirow,tabularx}
\usepackage{setspace}
\usepackage{romannum}
\title{Integrating Audio-Visual Features for Multimodal Deepfake Detection}
\name{Sneha Muppalla, Shan Jia, Siwei Lyu}%\thanks{Thanks to XYZ agency for funding.}}
\address{Author Affiliation(s)}

\name{Sneha Muppalla$^\star$, Shan Jia$^{\dagger\ast}$, and Siwei Lyu$^\dagger$}
\address{
$^\star$Cupertino High School, Cupertino, CA\\
$^\dagger$University at Buffalo, State University of New York\\
$^\ast${\small Corresponding Author}}
\begin{document}
%\ninept
\maketitle
\begin{abstract}
Deepfakes are AI-generated media in which an image or video has been digitally modified. The advancements made in deepfake technology have led to privacy and security issues. Most deepfake detection techniques rely on the detection of a single modality. Existing methods for audio-visual detection do not always surpass that of the analysis based on single modalities. Therefore, this paper proposes an audio-visual-based method for deepfake detection, which integrates fine-grained deepfake identification with binary classification. We categorize the samples into four types by combining labels specific to each single modality. This method enhances the detection under intra-domain and cross-domain testing.\let\thefootnote\relax\footnotetext{\textbf{979-8-3503-0965-2/23/\$31.00 ©2023 IEEE.}}
\end{abstract}
\section{Introduction}
\label{sec:intro}
Deepfakes are a type of synthetic media that alter and fabricate media in various ways, from images, videos, and audio recordings. They involve the use of generative adversarial networks (GANs) \cite{goodfellow2014generative}, Autoencoder~\cite{rumelhart1985learning}, Diffusion models~\cite{guarnera2023level} and other machine learning algorithms to be formed. The rapid development of such algorithms has made the process to create these synthetic videos increasingly easier and faster. The most well-known form of deepfakes involves generating a video in which one person's face is swapped onto another person's body. This creates a convincing illusion that the person whose face was used is the one acting in the video.
%placing someone’s face onto another persons body, creating highly realistic but fabricated content. 
This technology has found practical applications in various industries, particularly in the entertainment sector with film production and video gaming.

However, at the same time, deepfakes can be easily used for malicious purpose, posing a significant threat to media security and the integrity of information authentication.
%thus posing a harmful threat to society due to its harmful applications in various ways. 
%In law enforcement, there has been alteration of video evidence in criminal cases such as doorbell cameras and dialogue due to synthetic media technology. Courts and lawyers are struggling to grapple with the growing frequency of deepfakes, and this breaks the trust in the justice system \cite{Policechiefmagazine.org}. 
For example, financial frauds have occurred through the use of fake and stolen identities to open bank accounts. %There was one case where 
Bank robbers used an artificial intelligence cloning voice to steal 35 million from the UAE bank 
\footnote{https://gizmodo.com/bank-robbers-in-the-middle-east-reportedly-cloned-someo-1847863805} in 2021. Yet, the main concerns regarding deepfakes revolve around the spread of misinformation and the development of inappropriate content. In 2022, a deepfake video featuring Ukraine's President Volodymyr Zelensky announcing a surrender \footnote{https://www.youtube.com/watch?v=X17yrEV5sl4} was spread out to the public, causing widespread panic and confusion. %In 2022, a deepfake video was spread out in the public of Ukraine president Volodymyr Zelensky \cite{Allyn_2022b} announcing a surrender, leading to  sow panic and confusion.  %Such misinformation is another form of propaganda and false information that incites violence. There has also been an increasing form of inappropriate content and digital harassment that causes severe emotional distress \cite{Deepfake:} to the victims. 

%Existing detection strategies:
The great threat posed by deepfakes has prompted researchers to create effective deepfake detection methods in recent years~\cite{rana2022deepfake, jia2022model, sun2022faketracer, zhao2021multi}. Most methods use deep learning models to classify the input into ``real” or ``fake”, which is a binary classification. A majority of these methods target only a single modality \cite{weerawardana2021deepfakes}, mostly exploring the visual artifacts in videos. In the past few years, multimodal deepfakes, involving manipulations in either audio or video, have begun to emerge with a more realistic and diverse generation~\cite{khalid2021fakeavceleb, chen2022trusted}. Audio-visual detection methods~\cite{khalid2021evaluation, cozzolino2023audio,dolhansky2020deepfake,yang2023avoid} have been proposed to combine features from audio and video for deepfake detection. %There are many other applications of video processing that use and combine multiple modalities for visual-audio, literary, language, vision and emotion recognition. Through this implementation, 
The analysis of two modalities, such as by fusing audio-visual features~\cite{khalid2021fakeavceleb, khalid2021evaluation} or learning audio-visual inconsistencies~\cite{yang2023avoid, cozzolino2023audio, hashmi2022multimodal}, can provide rich information to expose the deepfakes. %When combined, the various modalities can be applied to determine whether a video or source is real or not. However, it is not always the case that multi-modals will result in a higher accuracy than applying a single modality. 
However, existing studies reveal that employing multimodalities does not guarantee a higher accuracy compared to the application of a single modality~\cite{khalid2021fakeavceleb, khalid2021evaluation, hashmi2022multimodal}. %There are cases where a single modality is better to use than multi modalities. 
%This can be influenced by factors such as the structure of the detection model,  the method used to combine multimodal features, and the selected dataset.
Exploring how to leverage multiple modalities to enhance deepfake detection deserves further research. %But in general, the fusion applied in multi modals provide a richer and inclusive experience, especially in terms of complex information like deepfake videos. 

Given that audio-visual deepfakes can be manipulated in various manners, such as pairing real video with synthesized audio, combining authentic audio with face-swapped video, or altering both modalities, it's important to consider these categories in exploring multi-modality features. However, existing feature fusion methods treat different types of deepfakes as the same fake type. This may confuse the feature learning of a single modality, especially in ensemble-based fusion~\cite{khalid2021fakeavceleb, khalid2021evaluation, hashmi2022multimodal}. For example, in the case of deepfakes with real video and fake audio, as well as deepfakes with fake video and fake audio, the labels of the samples are the same, yet the video branch might learn unstable features for fusion. Incorporating a single modality detection loss~\cite{hashmi2022multimodal} can alleviate the confusion. However, the fusion performance is not necessarily enhanced. We believe that constrained by the audio and video generation models, the inconsistencies in audio-visual artifacts vary among different types of deepfakes. Therefore, in this paper, instead of treating the audio-visual sample as a simple binary classification task as previous detection methods did, we propose to integrate a fine-grained deepfake identification module to guide the detection model to discern the distinct artifacts present in videos that are fake in a single modality or in both modalities.
%The purpose of this paper is to demonstrate the relationship between the visual and audio aspects of the same video. Prior methods have resulted in poor performance from the use of both single modality and multi-modalities. These include ensemble of two networks and multimodal-based detection\cite{khalid2021evaluation, zhao2021multi}. We envision a direction of deepfake detection methods that will help improve the results as the complexity of deepfake generation methods increase.
We present a simple yet effective approach to identify the visual-audio artifacts among four types of videos, including real video real audio, real video fake audio, fake video real audio, and fake video fake audio. Combined with the artifact learning from each single modality, we improve the performance of the audio-visual feature fusion method.   
We apply our approach to two detection backbones, including the widely-used Capsule network \cite{nguyen1910use} and the recent Swin Transformer \cite{liu2021swin} models. Experiments on two public multimodal deepfake datasets under intra-domain and cross-domain testing show superior performance of the proposed method to existing detectors.%, namely FakeAVCeleb \cite{khalid2021fakeavceleb} dataset and the TMC dataset \cite{chen2022trusted}. %. Using these features, we combine fine-grained Deepfake detection with single-modality binary classification. We train and test our method and compare the networks on the .  

%\subsubsection{Law enforcement}
%\label{sssec:subsubhead}

%\subsubsection{Inappropriate content}
%\label{sssec:subsubhead}

%\subsubsection{Cyberbullying}
%\label{sssec:subsubhead}

%\subsubsection{Misinformation}
%\label{sssec:subsubhead}

%\subsubsection{Financial Frauds}
%\label{sssec:subsubhead}
\vspace{-0.24cm}
\section{Related Work}
\vspace{-0.24cm}
\label{sec:format}

\subsection{Multimodal deepfake dataset}
\vspace{-0.24cm}
%As the domain of deepfakes began to get more awareness, researchers have started to create datasets that consist of real videos and deepfake videos. 
Several deepfake datasets have been established to advance the development of detection techniques. Early datasets focused on visual modality generation, either utilizing authentic audio or excluding audio entirely, such as %So it didn’t matter whether the audio was real or manipulated, or even if there was any audio at all in these datasets. 
%Some existing datasets that focus only on video manipulation 
FaceForensics++ (FF++) \cite{rossler2018faceforensics}, CelebDF~\cite{li2020celeb}, DFDC (Deepfake Detection Challenge) \cite{dolhansky2020deepfake}, and KoDF (Korean Deepfake Detection Dataset) \cite{kwon2021kodf}. %Other than the DFDC dataset, all the other datasets only include video clips with no audio. Although DFDC \cite{dolhansky2020deepfake} does include audio and the authors do claim that there is manipulation done on some of the audio clips, there is no file or any details that disclose where the manipulation occurs and is nowhere to be found. Furthermore, the ethnicity distribution is not clear in these datasets and can cause the models to me more biased towards certain ethnic backgrounds than others.
With the development of audio generation, audio-visual deepfake datasets combining the generation of cloned fake audio and generated faces have been created recently, including FakeAVCeleb dataset \cite{khalid2021fakeavceleb} and TMC~\cite{chen2022trusted}. These two datasets both cover fake video with real audio, real video with fake audio, fake video with fake audio, and real video with real audio categories. %s began to be produced through different techniques, so did the development of new datasets. 
%Multi-modal deepfakes dataset began to emerge that contain different variety of deepfakes that can be fake in multiple ways. Most of these types of datasets contain 3 different types of multi-modal deepfakes \cite{khalid2021fakeavceleb, chen2022trusted}: fakevideo-realaudio, realvideo-fakeaudio, fakevideo-fakeaudio. And then there are the real videos that have real video and real audio. Fake videos in these datasets are mainly created through 2 techniques: face swapping and lip syncing. There are hundreds of different algorithms that produce these two methods, but the final output is different from these algorithms and make such datasets more complex and better for methods to be tested and trained on. As for audio, human voice AI models are applied for audio manipulation in order to use fake audio in these datasets. Similar to the video, there are different algorithms and techniques to create fake audio that help make these datasets more challenging to identify the type of video.

\vspace{-0.24cm}
\subsection{Multimodal deepfake detection}
\vspace{-0.24cm}
\label{ssec:subhead}
%Due to the potential harms of deepfakes, 
There has been a surge of interest in deepfake detection. A majority of unimodal detection methods \cite{weerawardana2021deepfakes} focused on the visual and facial features. For multimodal deepfake detection, %deep learning and the application of deep neural networks are used. Deep 
deep learning-based audio-visual models have been proposed by researchers. %in which convolutional neural networks (CNN) \cite{zhang2016joint}, and recurrent neural networks (RNN) \cite{masi2020two} are combined. 
One branch of detection methods uses the fusion of features or scores from two modalities~\cite{khalid2021evaluation, cozzolino2023audio,hashmi2022multimodal}. However, the ensemble of audio and visual networks does not yield as impressive results as the detection methods that focus on a single modality. 
%In order to train for fake audio detection, feature-extraction occurs on the raw audio files in order to convert them into vectors to be applied in the model. The most common feature extraction methods are fast Fourier transforms and mel-frequency \cite{Roberts_2022}. As for fake video model training, the video is extracted frame by frame \cite{cozzolino2023audio} to provide more analysis and converted into vectors as well to be applied in the model. However, there hasn’t been a similar amount of focus regarding the use of multimodalities part of the same video. Current multimodal deepfake detection methods involve the use of different sorts of information, including visual, audio, and textual, in order to classify deepfake content. Through the combination and application of multiple modalities, these methods help improve the accuracy and robustness of deepfake detection systems. 
Another branch of methods leverages multimodal detectors that extract the audio-visual inconsistencies in deepfakes~\cite{yang2023avoid, cheng2022voice, cozzolino2023audio,feng2023self}. These methods have limitations in either delivering superior performance~\cite{khalid2021fakeavceleb, khalid2021evaluation}, or offering computationally efficient detection.
%Multimodal learning techniques are involved in various areas of human activity, from action recognition, face recognition, and emotion recognition. One common technique is fusion-based analysis. This method combines information from different aspects, such as visual, audio, and context. Through this, the fusion technique is able to leverage the strength of each modality, leading to an increase in overall performance for detection. One type of fusion is feature fusion. The extracted features from each modality are combined in the same layer of the model, and then fused together by concatenating the features together. Another type of fusion is score fusion, with the same goal of the feature fusion and leverage the complementary information provided by the different modalities. Rather than concatenating the extracted features, they are averaged to create a unified decision. 
\vspace{-0.44cm}
\section{Methodology}
\vspace{-0.24cm}
\label{sec:pagestyle}

\begin{figure}[tb]\centering
\includegraphics[width=8.3cm]{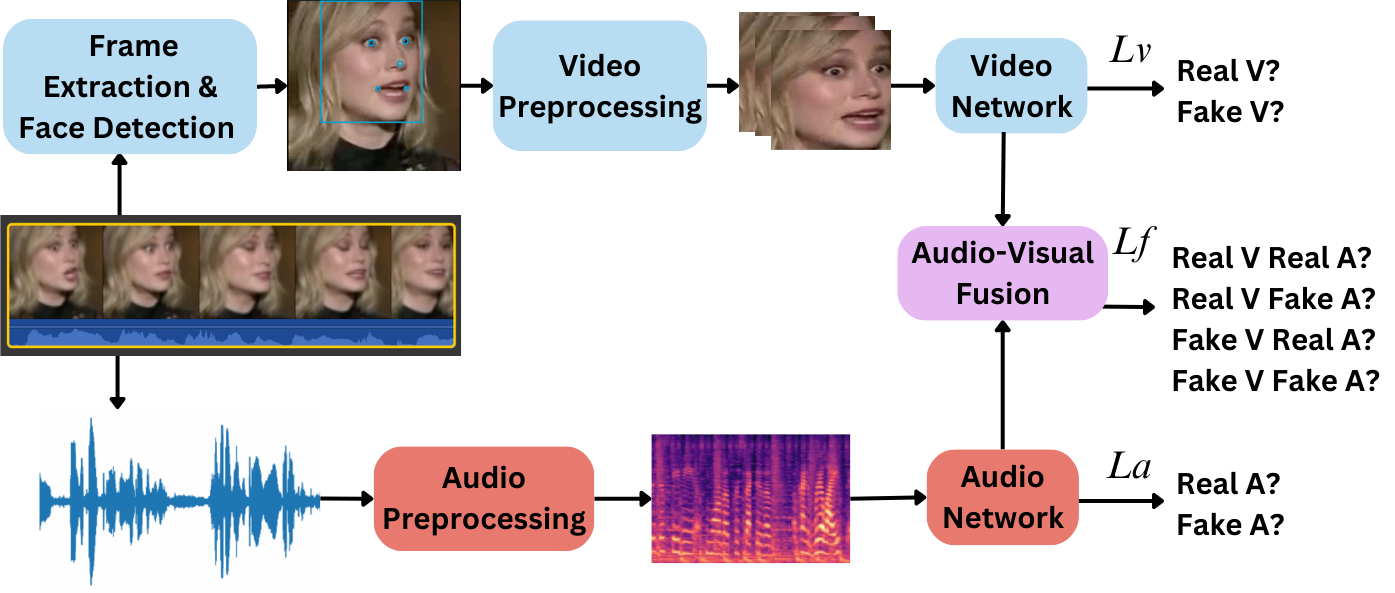}
\vspace{-0.34cm}
\caption{Pipeline of the proposed detection method.}
\vspace{-0.44cm}
\label{fig:res}
\end{figure}

\subsection{Overview}\vspace{-0.24cm}
%Our method involves frame-based classification and classify each frame as real or fake to give a combined final result for the classification of the video. 
We propose a simple yet effective detection method to improve the detection performance of multimodal deepfakes. The proposed method involves the fusion of audio and visual branches with a fine-grained deepfake classification loss. We further incorporate the binary classification losses from each modality to guide the detection model to capture both the inconsistency of multi-modality and the artifacts from each individual modality. In this section, we will first describe the preprocessing modules for each single modality, and then introduce the feature extraction networks, followed by a multi-task learning strategy. %evaluate the datasets through two different multimodels and explore multiple baselines to evaluate their effects on the FakeAVCeleb \cite{khalid2021fakeavceleb} and TMC dataset \cite{chen2022trusted}.
 \vspace{-0.24cm}
\subsection{Preprocessing}\vspace{-0.14cm}
%We first preprocess the datasets before passing it to the models for training and testing. Preprocessing was done separately for the video and audio aspects.
\subsubsection{Video Branch}\vspace{-0.14cm}
%The method for video preprocessing was the same for both the FakeAVCeleb dataset \cite{khalid2021fakeavceleb} and the TMC dataset \cite{chen2022trusted}. 
%Since the videos in FakeAVCeleb were originally from the VoxCeleb2 dataset \cite{chung2018voxceleb2}, the videos were already face-centered and cropped. However, there was still a lot of background behind the subject’s face in the videos. As for the TMC dataset, a lot of the videos contained the entire body of the subject, and there was a lot of excessive background. 
First, frame extraction is applied to each input video to output images. %The length of TMC videos were greater than a minute, so one frame was extracted for each second. The length for FakeAVCeleb videos were much shorter, at most 10 seconds, so almost every frame was extracted. 
Considering the varying lengths of video samples, we extract one frame per second for videos that are longer than a minute, such as the videos in TMC dataset~\cite{chen2022trusted}, and extract all frames for shorter videos, such as those in FakeAVCeleb~\cite{khalid2021fakeavceleb} dataset. 
For each frame, a face detection method based on MTCNN \cite{xiang2017joint} is used to detect and crop the face regions based on facial landmarks. %capture the dimensions and location of where the subject’s face is on the image. %MTCNN \cite{xiang2017joint}, or multi-task cascaded neural networks, is a neural network which detects faces and facial landmarks on images. 
%Using the location found by this tool, each frame was cropped such that only the person’s face remained in the final version. Also note that the frame had to be a square and not a rectangle, so the dimensions were of the shorter of the width and length so that no background would show in the final frame. 
\vspace{-0.24cm}
\subsubsection{Audio branch}\vspace{-0.14cm}
%Like the video, the preprocessing for audio was the same for both datasets. 
For the audio modality, we first extract the audio in a WAV format from the input video. The WAV file has the audio in a raw format, which cannot be passed into the detection model directly. Therefore, we further convert the audio into a mel-spectrogram image. A mel spectrogram \cite{Roberts_2022, sun2023ai} is a spectrogram where the frequencies are converted into the mel scale. To do this, the raw audio files are first used to take samples of air pressure over time in order to digitally represent an audio signal, thereby capturing a waveform for the signal. This converts the audio file into a digital representation of an audio signal. Then, the audio signals are mapped from the time domain to the frequency domain using the Fast Fourier Transform (FFT), an algorithm that can efficiently compute the Fourier transform, a mathematical formula that breaks down a signal into its individual frequencies and the frequency’s amplitude. This results in a spectrum. We then convert the frequency scale (y-axis) to a log scale and the color dimension (amplitude scale) to decibels to form the spectrogram. The y-axis is mapped onto the mel scale, through another mathematical operation, and this forms the mel spectrogram \cite{Roberts_2022}. The mel scale conversion equation is
\vspace{-0.12cm}
\begin{equation}
   Mel = \frac{(log(1+ (Hz/1000)))\times 1000}{log(2)}
\end{equation}
\vspace{-0.08cm}
where Hz is the frequency. We convert the file into a mel spectrogram instead of a spectrogram because the mel scale converts the frequencies such that equal distances in pitch sound equally distant to a listener. In order to make sure that the range in frequency and length shown for each mel spectrogram are the same, we set the frequency scale from 0 to 8000 Hz and the length to 4 seconds. %Videos that are less than 4 seconds will not be applied nor converted to a mel spectrogram.
\vspace{-0.14cm}
\subsection{Feature Extraction}\vspace{-0.14cm}
Feature extraction involves converting input images into high-level features, aiming to capture distinctive patterns within the data that are essential for detection tasks. The preprocessed visual face images and audio mel spectrograms are fed into deep neural networks to automatically extract features for deepfake detection. It is worth noting that our method does not impose any restrictions on the feature extraction models, and exhibits flexibility when applied across various network architectures. In our experiment, we will utilize the Capsule network~\cite{nguyen1910use}, widely used in deepfake detection methods, and the Swin Transformer~\cite{liu2021swin}, which has recently demonstrated powerful feature learning capabilities in various image classification tasks, as two examples for the video and audio networks to show the flexibility of our method.

\vspace{-0.24cm}
\subsection{Multi-task Learning}\vspace{-0.14cm}
As our method proposes the integration of a fine-grained deepfake identification module with the binary classification of each modality for distinct feature learning, we formulate a multi-task learning strategy by fusing three loss functions. The loss function is defined as,
\vspace{-0.12cm}
\begin{equation}
   L_{total} = L_{a} + L_{v} + L_{f} 
\end{equation}
\vspace{-0.12cm}
where \begin{math}L_{a}\end{math} and \begin{math}L_{v}\end{math} are the binary cross-entropy losses applied to the each single audio and video modality, respectively, while \begin{math}L_{f}\end{math} is a four-class cross-entropy  %(score of feature) 
loss that considers the different types of videos. % The purpose of adding  the fusion loss is to make up for the shortcomings of the single modalities and gives an improvement when both modalities are used in learning for classification of deepfake videos in the datasets. 
Each of the cross-entropy loss is defined as:
\vspace{-0.12cm}
\begin{equation}
   L = -(ylog(\hat{y}) + (1-y)log(1-\hat{y})
\end{equation}
%\vspace{-0.12cm}
where \begin{math}y\end{math} is the ground truth label and \begin{math}\hat{y}\end{math} is the predicted label. %The four-class predictions are converted to binary class labels before being inputted in the combined losses, where the deepfakes are the 3 different types of deepfake videos, and the real videos are real video and audio. This emphasizes the influence the fusion has on improving the classification of the videos \cite{cozzolino2023audio}. 

For the audio-visual module which is a multi-class classification task, we average the video network output of multiple frames to represent the overall features of the input video. Two versions of audio-visual fusion can be implemented. The first involves feature fusion, where the average visual features from the video network are concatenated with the audio features from the audio network (after removing the classification head). The second approach is score fusion, which averages the scores from the video and audio networks that include classification heads. Subsequently, either the fused feature or the fused score will be fed into the four-class classification head for video identification. % If the average value is less than 0.5, the visual part of the video is determined as real, otherwise it is determined as fake if greater than or equal to 0.5. The same approach is applied for the audio spectrograms classifications for each video. 

\vspace{-0.24cm}
\section{Experiments}\vspace{-0.24cm}
\label{sec:typestyle}

\subsection{Implementation Details}
\vspace{-0.24cm}
%The multi-task strategy was applied to both a multi-modal capsule network \cite{nguyen1910use} and a transformer \cite{liu2021swin}. This section will go over the architecture of both of these models, the test settings, and the two different evaluation metrics used to compare the accuracies over multiple testings.
The frames extracted from videos were resized to 300\begin{math}\times\end{math}300, and the image of the audio mel-spectrogram was also resized to 300\begin{math}\times\end{math}300.
The detection models were trained for 50 epochs, with a batch size of 10. We employed the Adam optimizer \cite{kingma2014adam} with \begin{math}\beta_{1} = 0.9\end{math}, \begin{math}\beta_{2} = 0.999\end{math}, and a learning rate of \begin{math}5 \times 10^{-4}.\end{math} 

For comparison with other methods, we report two widely-used metrics in deepfake detection, namely Accuracy and Area Under Curve (AUC) which is a metric that quantifies the total area beneath the curve generated when plotting the True Positive Rate against the False Positive Rate.

\vspace{-0.14cm}
\subsection{Dataset}\vspace{-0.14cm}
We evaluated the performance of our method on two multimodal deepfake datasets, namely FakeAVCeleb \cite{khalid2021fakeavceleb} and TMC dataset \cite{chen2022trusted} using the 80-20 training/testing split. 

\textbf{FakeAVCeleb} \cite{khalid2021fakeavceleb} dataset was generated using videos from VoxCeleb2 \cite{chung2018voxceleb2} which consists of real Youtube videos of celebrities with five ethnic backgrounds: Caucasian (Americans), Caucasian (Europeans), African American, South Asian, and East Asian. There is also equal distribution between males and females for each of the ethnic backgrounds. Each ethnic group consists of 100 real videos of 100 celebrities, 50 for each gender. This outputs 600 different videos with an average duration of about 7 seconds. The dataset consists of three different types of audio-video deepfakes (fake audio only, fake video only, or both) using popular deepfake generation methods. These methods include face-swapping and facial reenactment methods. For fake audio generation, they used synthetic speech synthesis methods to generate cloned or fake voice samples. Wav2Lip was used for facial reenactment based on the audio source.

\textbf{TMC Dataset} \cite{chen2022trusted}, created in 2022, was designed to tackle the issue of fake media in Singapore. This dataset consists of 4,380 fake videos and 2,563 real-life footage from several different sources, including presenters and journalists in news programs, interviews answering questions on live TV, and people talking about general topics. %Compared to the FakeAVCeleb \cite{khalid2021fakeavceleb} dataset, 
TMC contains 72.65\% of subjects Asians, and 45.82\% of female subjects. The duration of these videos varies from 10 seconds to a minute. There are 4 different types of fakes: fake audio only, fake video only, both aspects are fake, or both are real but the audio and video do not match. Similar to FakeAVCeleb, TMC was generated using popular deepfake generation methods, including face-swapping and deep learning techniques. StarGAN-VC was used for fake audio generation. 

\subsection{Baselines}\vspace{-0.14cm}
We first compare the proposed audio-visual method with different detection strategies under two feature learning backbones, namely the Capsule network \cite{nguyen1910use} and the Swin Transformer \cite{liu2021swin} model. %These detection strategies include unimodals using video or audio modality, audio-visual feature fusion and score fusion 
Then we conduct a comparison with existing deepfake detection methods \cite{khalid2021evaluation, cheng2022voice} that were evaluated on the FakeAVCeleb dataset, %As of right now, no existing methods have applied the TMC dataset \cite{chen2022trusted} nor these two datasets through cross testing. For the different losses, we test the models only on the audio loss, only on the video loss, binary loss feature fusion, and the binary loss score fusion. We also compare the results with publicly available implementations \cite{khalid2021evaluation, cheng2022voice} on the FakeAVCeleb dataset,
including ensemble frame-based methods of MesoInception-4 \cite{afchar2018mesonet}, EfficientNet \cite{tan2019efficientnet}, and FTCN \cite{zheng2021exploring}, and audio-visual multimodal methods including AVoiD-DF \cite{yang2023avoid} and AV-Lip-Sync \cite{shahzad2022lip}. %The results from these models are not given based on modality, but rather overall detection.

\begin{center}
\begin{table}[!t]
\footnotesize
\setlength\tabcolsep{3pt}
\begin{tabularx}{0.485\textwidth}{c|c|c|cc|cc|cl}
\hline
Dataset&Model& Method & \multicolumn{2}{c|}{Overall} & \multicolumn{2}{c|}{Video} & \multicolumn{2}{c}{Audio}\\
&&&AUC&ACC&AUC&ACC&AUC&ACC\\
\hline
\hline
\multirow{11}{0.8cm}{FakeAV- Celeb to FakeAV- Celeb} 
                &\multirow{5}{1.1cm}{Capsule Forensics}  & Ours-S  & 99.30 & 99.20 & \textbf{97.27} & 96.43 & \textbf{99.75} & \textbf{99.80}\\
                  %&& Video & 81.3 & 80.6  & 96.53    & \textbf{96.51} & 66.71 & 66.5\\
                  %&& Audio & 68.45 & 68.36 & 40.27 & 40.14 & 99.11& 99.06       \\
                  && Video & - & -  & 96.53    & \textbf{96.51} & - & -\\
                  && Audio & - & - & - & - & 99.11& 99.06       \\
                  && Feature    & \textbf{99.44} & 99.16 & 26.74 & 26.78 & 55.61 & 55.57 \\
                  && Score & 99.25 & \textbf{99.75} & 75.48 & 75.41 & 77.81 & 77.83 \\\cline{2-9}
&\multirow{5}{1.1cm}{Swin Tr-ansformer}  & Ours-S  & \textbf{91.65} & \textbf{90.51} & \textbf{88.13} & \textbf{96.21} & 93.47 & \textbf{94.79}\\
                && Ours-F & 85.67 & 86.73 & 79.23 & 86.77 & \textbf{94.57} & 91.26 \\
                  %&& Video & 88.91 & 88.14  & 86.54& 86.56 & 87.78 & 86.56\\
                  %&& Audio & 85.97 & 87.92 & 72.61 & 72.64 & 88.06 & 88.06\\
                  && Video & -& -  & 86.54& 86.56 & -& -\\
                  && Audio & - & - & - & - & 88.06 & 88.06\\
                  && Feature & 82.91 & 89.68 & 81.33 & 80.26 & 90.28 &92.16 \\
                  && Score & 87.91 & 90.55 & 89.11 & 90.29 & 90.13 & 94.22 \\
\hline
\multirow{10}{1cm}{TMC to TMC} & \multirow{5}{1.1cm}{Capsule Forensics}  & Ours-S  & \textbf{93.67} & \textbf{93.15} & \textbf{92.03} & \textbf{91.98} & \textbf{99.73} & \textbf{99.68}\\
                  %&& Video & 77.53 & 78.46  & 90.97& 90.91 & 60.7 & 62.75\\
                  %&& Audio & 75.83 & 75.66 & 56.21 & 57.19 & 99.07 & 98.72       \\
                  && Video & - & -  & 90.97& 90.91 & - & -\\
                  && Audio & - & - & -& - & 99.07 & 98.72       \\
                  && Feature   & 88.89 & 89.24 & 50.57 & 51.79 & 29.91 & 28.85 \\
                  && Score  & 83.00 & 84.36 & 75.16 & 75.61 & 78.30 & 76.54 \\\cline{2-9}
&\multirow{5}{1.1cm}{Swin Tr-ansformer} & Ours-S  & \textbf{80.11} & \textbf{85.57} & \textbf{72.39} & \textbf{73.02} & \textbf{83.10} & \textbf{87.29}\\
             && Ours-F & 65.79 & 78.23 & 70.55 & 65.35 & 72.73 & 79.33 \\
                  %&& Video & 52.23 & 71.75  & 50.72& 54.07 & 51.83 & 52.46\\
                  %&& Audio & 70.61 & 70.46 & 53.68 & 52.76 & 72.63 & 76.81\\
                  && Video & - & -  & 50.72& 54.07 & - & -\\
                  && Audio & - & - & - & - & 72.63 & 76.81\\
                  && Feature & 67.28 & 72.87 & 57.47 & 62.22 & 72.93 & 73.21 \\
                  && Score & 75.41 & 78.77 & 68.84 & 72.04 & 81.30 & 80.59 \\
\hline
\end{tabularx}\vspace{-0.24cm}
\caption{Comparison results under intra-domain testing.}
\label{tab:1}
\vspace{-0.3cm}
\end{table}
\end{center}

\begin{center}
\begin{table}\centering
\footnotesize
\begin{tabular}{c|cc}
\hline
\hline
Model& \multicolumn{2}{c}{Overall}\\
&AUC&ACC\\\cline{1-3}
MesoInception-4 \cite{afchar2018mesonet} & 72.22  & 75.82\\
FTCN \cite{zheng2021exploring} &84.00&64.90\\
EfficientNet \cite{tan2019efficientnet} & 81.03 & -\\
AVoiD-DF \cite{yang2023avoid} & 89.20 & 83.70\\
AV-Lip-Sync \cite{shahzad2022lip} & - & 94.00\\
Ours-S-Capsule Forensics  & \textbf{99.30} & \textbf{99.20}\\
Ours-S-Swin Transformer & \textbf{91.65} & \textbf{90.51}\\
\hline   
\end{tabular}\vspace{-0.24cm}
\caption{Comparison results with existing deepfake detection methods using reported results on FakeAVCeleb dataset.}%with single modality models Xception, F\begin{math}^{3}\end{math}-Net, and ensemble frame-based methods MesoInception-4, EfficientNet, and FTCN on FakeAVCeleb dataset}
\vspace{-0.3cm}
\label{tab:t2}
\end{table}
\end{center}

\vspace{-2.2cm}
\subsection{Intra-domain testing}\vspace{-0.14cm}
%There were two different versions of intradataset testing done on the models: one on the FakeAVCeleb dataset \cite{khalid2021fakeavceleb} and the other on the TMC dataset \cite{chen2022trusted}. 
%Starting with the FakeAVCeleb dataset, each type of deepfake consisted of 5000 frames each, with 5 frames from 1000 selected videos. Additional videos were taken from the VoxCeleb2 \cite{chung2018voxceleb2} dataset to be part of the real section due to the number of real videos in the FakeAVCeleb dataset being limited to 500. As for testing, 5 frames were selected out of 250 randomly selected videos, and this was the same for all types of videos, including the different versions of deepfakes as well as real videos. The same process was done for audio, with the same audio mel-spectrogram \cite{Roberts_2022} being used for frames part of the same video. A similar technique was done for the intra-testing on the TMC Dataset. 250 videos were selected from the dataset for training, in which 20 frames were added to the training set for both real videos and deepfakes. For testing, 20 frames were chosen from 70 other videos not part of the training set. 
Under intra-domain testing scenarios, the detection model is trained and tested in the same dataset. Table \ref{tab:1} compares our score fusion method ``Ours-S" and feature fusion method ``Ours-F" with two single modality models (``Video" and ``Audio") and traditional feature and score fusion methods based on a binary classification loss. With the advantage of the fine-grained video identification module, our method, especially with score fusion, clearly surpasses other approaches, achieving the highest AUC and ACC scores in the majority of cases. This shows the effectiveness of the proposed method in identifying the artifacts from multi-modalities. It is worth noting that all methods achieve higher results on the FakeAVCeleb dataset than on the TMC dataset. This can be attributed to the high diversity and additional perturbations added to both video and
audio tracks in TMC dataset. %It is worth noting that the feature fusion methods tend to perform worse than the score fusion in both our strategy and traditional methods. One potential reason is that the
We further compare our score-fusion-based method with existing deepfake detection techniques on the FakeAVCeleb dataset. The results in Table \ref{tab:t2} show the outstanding performance of the proposed method in detecting multimodal deepfakes, especially using the Capsule network.

\vspace{-0.53cm}
\subsection{Cross-domain testing}\vspace{-0.14cm}
To evaluate the generalization ability of the proposed method in detecting unseen deepfakes, the proposed method is trained and tested using different datasets. %For cross-testing, the first evaluation for the models was training on the FakeAVCeleb \cite{khalid2021fakeavceleb} dataset. Each type of deepfake consisted of 5000 frames for training, in which 50 frames were chosen out of 100 selected videos. For the real videos, additional videos were taken from the VoxCeleb2 dataset \cite{chung2018voxceleb2} so there were 5000 frames as well in the real videos, but 5 frames for each of 1000 selected real videos selected. This expanded the range of various videos to be applied for the model to train on. To test on the TMC dataset \cite{chen2022trusted}, 12500 frames were applies with 5 frames chosen each from 250 of selected videos of each type of video and deepfakes in the dataset. The same approach is done on the cross-dataset testing for training on TMC \cite{chen2022trusted} and testing on FakeAVCeleb \cite{khalid2021fakeavceleb}.
The comparison results in Table \ref{tab:t3} show that our proposed method achieved the best performance under both cross-domain testing scenarios. We can also observe that training on TMC and testing on the FakeAvCeleb dataset leads to higher AUC and ACC for most cases. This is reasonable as the model trained on a more diverse dataset (i.e., TMC) would exhibit greater generalizability to unseen data. Moreover, it can be seen that the Swin Transformer obviously outperforms the Capsule network, demonstrating better generalization ability.

\begin{table}[!t]
\footnotesize
\setlength\tabcolsep{3pt}
\begin{tabularx}{0.49\textwidth}{c|c|c|cc|cc|cl}
\hline
Dataset&Model& Method& \multicolumn{2}{c}{Overall} & \multicolumn{2}{c|}{Video} & \multicolumn{2}{c}{Audio}\\
&&&AUC&ACC&AUC&ACC&AUC&ACC\\
\hline
\hline
\multirow{10}{0.8cm}{FakeAV- Celeb to TMC} & \multirow{5}{1.1cm}{Capsule Forensics}  & Ours-S  & \textbf{65.28} & \textbf{67.39} & \textbf{73.49} & \textbf{72.82} & \textbf{61.04} & \textbf{65.66}\\
                  && Video  & - & -  & 70.62    & 69.75 & - & -\\
                  && Audio & - & -& - & - & 53.74& 55.65       \\
                  && Feature   & 61.59 & 59.83 & 59.98 & 58.97 & 50.00 & 52.22 \\
                  && Score  & 60.67 & 65.11 & 63.89 & 63.48 & 50.03 & 47.84 \\\cline{2-9}
&\multirow{7}{1.1cm}{Swin Tr-ansformer}  & Ours-S   & \textbf{70.41} & \textbf{79.68} & \textbf{67.34} & \textbf{68.89} & \textbf{70.19} & \textbf{71.54}\\
                   % && Ours-F  & 60.26 & 69.49 & 51.39 & 65.94 & 47.09 & \textbf{73.67}\\
                  && Video & - & - & 55.41 & 56.72 & - & -\\
                  && Audio & - & - & - & - & 65.04 & 64.68\\
                  && Feature & 63.79 & 77.83 & 63.28 & 62.67 & 66.80 &68.23 \\
                  && Score  & 64.82 & 76.23 & 66.83 & 68.19 & 69.25 & 69.23 \\
                  %&& Naive Label Fusion & - & 67.91 & - & 56.77 & - & 69.25\\
\hline
\multirow{11}{0.8cm}{TMC to FakeAV- Celeb} & \multirow{5}{1.1cm}{Capsule Forensics}  & Ours-S  & \textbf{77.89} & \textbf{81.12} & \textbf{70.29} & \textbf{77.96} & \textbf{69.33} & \textbf{71.65}\\
                  && Video & - & - & 65.59& 62.24 & - & -\\
                  && Audio & - & -& - & - & 51.00 & 71.59       \\
                  && Feature   & 64.60 & 66.87 & 49.00 & 28.55 & 60.34 & 63.87 \\
                  && Score  & 66.41 & 66.77 & 64.84 & 72.04 & 65.30 & 63.59 \\\cline{2-9}
&\multirow{6}{1.1cm}{Swin Tr-ansformer}  & Ours-S  & \textbf{82.17} & \textbf{89.73} & \textbf{78.77} & \textbf{78.89} & \textbf{87.55} & \textbf{88.32}\\
                %&& Ours-F  & 74.11 & 78.23 & 63.49 & 77.57 & 80.31 & 80.31 \\
                  && Video & - & -  & 72.77 & 72.11 & - & -\\
                  && Audio & - & -& - & - & 78.09 & 82.45\\
                  && Feature & 76.10 & 84.17 & 71.97 & 74.23 & 80.48 & 85.91 \\
                  && Score & 80.56 & 87.88 & 73.76 & 77.69 & 85.47 & 85.34 \\
                  \hline
\end{tabularx}\vspace{-0.24cm}
\caption{Comparison results under cross-domain testing.}
\label{tab:t3}
\vspace{-0.15cm}
\end{table}

\begin{table}\centering
\footnotesize
\begin{tabular}{c|p{1cm}|cc}
\hline
Dataset& Method & \multicolumn{2}{|c}{Overall}\\
&&AUC&ACC\\
\hline
\hline
TMC to TMC & Ours-S & 82.18 & 75.29\\
FakeAVCeleb to TMC & Ours-S & 63.83 & 69.63\\
FakeAVCeleb to TMC & Naive & - & 44.00\\
\hline                 
\end{tabular}
\caption{Comparison results using Swin Transformer under cross-type testing.}
\vspace{-0.14cm}
\label{tab:t4}
\end{table}
\vspace{-0.34cm}
\subsection{Cross-type testing}\vspace{-0.14cm}
There is one type of cheap fake videos in the TMC dataset, namely real videos with mismatched real audio. To further demonstrate the effectiveness of our method in capturing audio-visual consistency features, we use the pre-trained models to test these mismatched videos as cross-type testing. Table \ref{tab:t4} compares our score-fusion based Swin Transformer model with the naive fusion method which uses the maximum predictions of two modalities. It is evident that our method is capable of handling these previously unseen fake types, while the naive fusion method fails. We also find that the majority of mismatched real video and audio samples are categorized under the ``Real Video Fake Audio" class within our multi-class identification module. This can be attributed to the similar audio-visual patterns, such as lack of synchronization, found in these two types of fake videos.
%\subsection{Visualization}

%\begin{figure}[tb]
% \centering
%  \centerline{\includegraphics[width=7.5cm]{visualization.png}}
%  \vspace{1.5cm}
%
%\caption{Heatmap results of each type of deepfake video, aswell as real videos}
%\label{fig:res}
%
%\end{figure}
%Refer to Grad-CAM activation map:   
% Below is an example of how to insert images. Delete the ``\vspace'' line,
% uncomment the preceding line ``\centerline...'' and replace ``imageX.ps''
% with a suitable PostScript file name.
% -------------------------------------------------------------------------

\section{Conclusion}
We introduce a method for audio-visual deepfake detection that identifies the multimodal inconsistency features across various deepfake types, as well as artifacts within each modality. The proposed method demonstrates good adaptability, allowing it to be applied across various feature extraction networks. Experimental results on two audio-visual deepfake datasets, both within-domain and cross-domain, highlight the effectiveness of the method. In our future work, we plan to focus on developing more robust multimodal networks to enhance the feature learning and fusion strategies for audio and video modalities. Additionally, exploring methods to improve the generalizability of multimodal deepfake detection methods to unseen deepfakes will be a key area of investigation.
%Our proposed multi-task strategy and method can be applied to synthetic media detection and digital images, including computer generated sources. 
%It can be shown that the capsule network yields a higher result when intradataset occurs or when similar methods to generate deepfakes are applied in training and testing. When methods and sources are quite different in training and testing, the transformer results in a higher accuracy. This demonstrates the multitask strategy is able to be effective in various methods in different training and testing modes. Overall, both models when using our strategy returned higher accuracies and metrics compared to previous methods tested. 
%These results and the detailed analysis gained from this research can help lead to further research and development on the detection on deepfakes, specifically on multi-modality related deepfakes. This can also be applied in areas outside of digital forensics as well. Future work could include the impact of preprocessing methods that affect the results, application of other architectures, and detecting if a certain section of a video or other source is fake.

\label{sec:majhead}
\footnotesize
\begin{spacing}{0.12}
\bibliographystyle{IEEEbib}
\bibliography{refs}

\begin{thebibliography}{10}

\bibitem{goodfellow2014generative}
Ian Goodfellow, Jean Pouget-Abadie, et~al.,
\newblock ``Generative adversarial nets,''
\newblock {\em Advances in neural information processing systems}, vol. 27,
  2014.

\bibitem{rumelhart1985learning}
David~E Rumelhart, Geoffrey~E Hinton, Ronald~J Williams, et~al.,
\newblock ``Learning internal representations by error propagation,'' 1985.

\bibitem{guarnera2023level}
Luca Guarnera, Oliver Giudice, and Sebastiano Battiato,
\newblock ``Level up the deepfake detection: a method to effectively
  discriminate images generated by gan architectures and diffusion models,''
\newblock {\em arXiv preprint arXiv:2303.00608}, 2023.

\bibitem{rana2022deepfake}
Md~Shohel Rana, Mohammad~Nur Nobi, Beddhu Murali, and Andrew~H Sung,
\newblock ``Deepfake detection: A systematic literature review,''
\newblock {\em IEEE access}, vol. 10, pp. 25494--25513, 2022.

\bibitem{jia2022model}
Shan Jia, Xin Li, and Siwei Lyu,
\newblock ``Model attribution of face-swap deepfake videos,''
\newblock in {\em 2022 IEEE International Conference on Image Processing
  (ICIP)}. IEEE, 2022, pp. 2356--2360.

\bibitem{sun2022faketracer}
Pu~Sun, Yuezun Li, Honggang Qi, and Siwei Lyu,
\newblock ``Faketracer: Exposing deepfakes with training data contamination,''
\newblock in {\em 2022 IEEE International Conference on Image Processing
  (ICIP)}. IEEE, 2022, pp. 1161--1165.

\bibitem{zhao2021multi}
Hanqing Zhao, Wenbo Zhou, Dongdong Chen, Tianyi Wei, Weiming Zhang, and Nenghai
  Yu,
\newblock ``Multi-attentional deepfake detection,''
\newblock in {\em Proceedings of the IEEE/CVF conference on computer vision and
  pattern recognition}, 2021, pp. 2185--2194.

\bibitem{weerawardana2021deepfakes}
MC~Weerawardana and TGI Fernando,
\newblock ``Deepfakes detection methods: A literature survey,''
\newblock in {\em 2021 10th International Conference on Information and
  Automation for Sustainability}. IEEE, 2021, pp. 76--81.

\bibitem{khalid2021fakeavceleb}
Hasam Khalid, Shahroz Tariq, et~al.,
\newblock ``Fakeavceleb: A novel audio-video multimodal deepfake dataset,''
\newblock {\em arXiv preprint arXiv:2108.05080}, 2021.

\bibitem{chen2022trusted}
Weiling Chen, Sheng Lun~Benjamin Chua, Stefan Winkler, and See-Kiong Ng,
\newblock ``Trusted media challenge dataset and user study,''
\newblock in {\em Proceedings of the 31st ACM International Conference on
  Information \& Knowledge Management}, 2022, pp. 3873--3877.

\bibitem{khalid2021evaluation}
Hasam Khalid, Minha Kim, et~al.,
\newblock ``Evaluation of an audio-video multimodal deepfake dataset using
  unimodal and multimodal detectors,''
\newblock in {\em Proceedings of the 1st workshop on synthetic
  multimedia-audiovisual deepfake generation and detection}, 2021, pp. 7--15.

\bibitem{cozzolino2023audio}
Davide Cozzolino, Alessandro Pianese, Matthias Nie{\ss}ner, and Luisa
  Verdoliva,
\newblock ``Audio-visual person-of-interest deepfake detection,''
\newblock in {\em Proceedings of the IEEE/CVF Conference on Computer Vision and
  Pattern Recognition}, 2023, pp. 943--952.

\bibitem{dolhansky2020deepfake}
Brian Dolhansky, Joanna Bitton, et~al.,
\newblock ``The deepfake detection challenge (dfdc) dataset,'' 2020.

\bibitem{yang2023avoid}
Wenyuan Yang, Xiaoyu Zhou, Zhikai Chen, et~al.,
\newblock ``Avoid-df: Audio-visual joint learning for detecting deepfake,''
\newblock {\em IEEE Transactions on Information Forensics and Security}, vol.
  18, pp. 2015--2029, 2023.

\bibitem{hashmi2022multimodal}
Ammarah Hashmi, Sahibzada~Adil Shahzad, et~al.,
\newblock ``Multimodal forgery detection using ensemble learning,''
\newblock in {\em 2022 Asia-Pacific Signal and Information Processing
  Association Annual Summit and Conference (APSIPA ASC)}. IEEE, 2022, pp.
  1524--1532.

\bibitem{nguyen1910use}
HH~Nguyen, J~Yamagishi, and I~Echizen,
\newblock ``Use of a capsule network to detect fake images and videos. arxiv
  2019,''
\newblock {\em arXiv preprint arXiv:1910.12467}.

\bibitem{liu2021swin}
Ze~Liu, Yutong Lin, Yue Cao, Han Hu, Yixuan Wei, Zheng Zhang, Stephen Lin, and
  Baining Guo,
\newblock ``Swin transformer: Hierarchical vision transformer using shifted
  windows,''
\newblock in {\em Proceedings of the IEEE/CVF international conference on
  computer vision}, 2021, pp. 10012--10022.

\bibitem{rossler2018faceforensics}
Andreas R{\"o}ssler, Davide Cozzolino, et~al.,
\newblock ``Faceforensics: A large-scale video dataset for forgery detection in
  human faces,''
\newblock {\em arXiv preprint arXiv:1803.09179}, 2018.

\bibitem{li2020celeb}
Yuezun Li, Xin Yang, Pu~Sun, Honggang Qi, and Siwei Lyu,
\newblock ``Celeb-df: A large-scale challenging dataset for deepfake
  forensics,''
\newblock in {\em Proceedings of the IEEE/CVF conference on computer vision and
  pattern recognition}, 2020, pp. 3207--3216.

\bibitem{kwon2021kodf}
Patrick Kwon, Jaeseong You, Gyuhyeon Nam, Sungwoo Park, and Gyeongsu Chae,
\newblock ``Kodf: A large-scale korean deepfake detection dataset,''
\newblock in {\em Proceedings of the IEEE/CVF International Conference on
  Computer Vision}, 2021, pp. 10744--10753.

\bibitem{cheng2022voice}
Harry Cheng, Yangyang Guo, et~al.,
\newblock ``Voice-face homogeneity tells deepfake,''
\newblock {\em arXiv preprint arXiv:2203.02195}, 2022.

\bibitem{feng2023self}
Chao Feng, Ziyang Chen, and Andrew Owens,
\newblock ``Self-supervised video forensics by audio-visual anomaly
  detection,''
\newblock in {\em Proceedings of the IEEE/CVF Conference on Computer Vision and
  Pattern Recognition}, 2023, pp. 10491--10503.

\bibitem{xiang2017joint}
Jia Xiang and Gengming Zhu,
\newblock ``Joint face detection and facial expression recognition with
  mtcnn,''
\newblock in {\em 2017 4th international conference on information science and
  control engineering (ICISCE)}. IEEE, 2017, pp. 424--427.

\bibitem{Roberts_2022}
Leland Roberts,
\newblock ``Understanding the mel spectrogram,'' Aug 2022.

\bibitem{sun2023ai}
Chengzhe Sun, Shan Jia, Shuwei Hou, and Siwei Lyu,
\newblock ``Ai-synthesized voice detection using neural vocoder artifacts,''
\newblock in {\em Proceedings of the IEEE/CVF Conference on Computer Vision and
  Pattern Recognition}, 2023, pp. 904--912.

\bibitem{kingma2014adam}
Diederik~P Kingma and Jimmy Ba,
\newblock ``Adam: A method for stochastic optimization,''
\newblock {\em arXiv preprint arXiv:1412.6980}, 2014.

\bibitem{chung2018voxceleb2}
Joon~Son Chung, Arsha Nagrani, and Andrew Zisserman,
\newblock ``Voxceleb2: Deep speaker recognition,''
\newblock {\em arXiv preprint arXiv:1806.05622}, 2018.

\bibitem{afchar2018mesonet}
Darius Afchar, Vincent Nozick, Junichi Yamagishi, and Isao Echizen,
\newblock ``Mesonet: a compact facial video forgery detection network,''
\newblock in {\em 2018 IEEE international workshop on information forensics and
  security (WIFS)}. IEEE, 2018, pp. 1--7.

\bibitem{tan2019efficientnet}
Mingxing Tan and Quoc Le,
\newblock ``Efficientnet: Rethinking model scaling for convolutional neural
  networks,''
\newblock in {\em International conference on machine learning}. PMLR, 2019,
  pp. 6105--6114.

\bibitem{zheng2021exploring}
Yinglin Zheng, Jianmin Bao, Dong Chen, Ming Zeng, and Fang Wen,
\newblock ``Exploring temporal coherence for more general video face forgery
  detection,''
\newblock in {\em Proceedings of the IEEE/CVF international conference on
  computer vision}, 2021, pp. 15044--15054.

\bibitem{shahzad2022lip}
Sahibzada~Adil Shahzad, Ammarah Hashmi, et~al.,
\newblock ``Lip sync matters: A novel multimodal forgery detector,''
\newblock in {\em 2022 Asia-Pacific Signal and Information Processing
  Association Annual Summit and Conference (APSIPA ASC)}. IEEE, 2022, pp.
  1885--1892.

\end{thebibliography}
\end{spacing} 
\end{document}